\documentclass[journal]{IEEEtran}

\usepackage{cite}
\usepackage{amsmath,amssymb,amsfonts}
\usepackage{algorithmic}
\usepackage{graphicx}
\usepackage{textcomp}
\usepackage{xcolor}
\usepackage{booktabs}
\usepackage{multirow}
\usepackage{hyperref}
\usepackage{subcaption}
\usepackage{balance}

\graphicspath{{figures/}}

\begin{document}

\title{When Does Multimodal AI Help? Diagnostic Complementarity of Vision-Language Models and CNNs for Spectrum Management in Satellite-Terrestrial Networks}

\author{\IEEEauthorblockN{Yuanhang Li}}

\maketitle

\begin{abstract}
The adoption of vision-language models (VLMs) for wireless network management is accelerating, yet no systematic understanding exists of where these large foundation models outperform lightweight convolutional neural networks (CNNs) for spectrum-related tasks. This paper presents the first diagnostic comparison of VLMs and CNNs for spectrum heatmap understanding in non-terrestrial network and terrestrial network (NTN-TN) cooperative systems. We introduce SpectrumQA, a benchmark comprising 108K visual question-answer pairs across four granularity levels: scene classification (L1), regional reasoning (L2), spatial localization (L3), and semantic reasoning (L4). Our experiments on three NTN-TN scenarios with a frozen Qwen2-VL-7B and a trained ResNet-18 reveal a clear task-dependent complementarity: CNN achieves 72.9\% accuracy at severity classification (L1) and 0.552 IoU at spatial localization (L3), while VLM uniquely enables semantic reasoning (L4) with F1=0.576 using only three in-context examples---a capability fundamentally absent in CNN architectures. Chain-of-thought (CoT) prompting further improves VLM reasoning by 12.6\% (F1: 0.209$\to$0.233) while having zero effect on spatial tasks, confirming that the complementarity is rooted in architectural differences rather than prompting limitations. A deterministic task-type router that delegates supervised tasks to CNN and reasoning tasks to VLM achieves a composite score of 0.616, a 39.1\% improvement over CNN alone. We further show that VLM representations exhibit stronger cross-scenario robustness, with smaller performance degradation in 5 out of 6 transfer directions. These findings provide actionable guidelines: deploy CNNs for spatial localization and VLMs for semantic spectrum reasoning, rather than treating them as substitutes.
\end{abstract}

\begin{IEEEkeywords}
Vision-language models, spectrum management, satellite-terrestrial networks, visual question answering, diagnostic evaluation, task complementarity
\end{IEEEkeywords}

\section{Introduction}
\label{sec:intro}

The integration of artificial intelligence into wireless network management has progressed rapidly, from deep reinforcement learning (DRL) for resource allocation~\cite{ledsm2025,lamdrl2025} to large language model (LLM)-based network agents~\cite{wirelessllm2024}. Most recently, vision-language models (VLMs)---foundation models capable of jointly processing images and text---have emerged as candidates for radio frequency (RF) signal understanding. Works such as ``Seeing Radio''~\cite{seeingradio2026} demonstrate that VLMs can classify modulation types from spectrograms, while RF-GPT~\cite{rfgpt2026} extends this to multi-task RF reasoning.

However, a fundamental question remains unanswered: \textit{when do VLMs actually outperform lightweight CNNs for spectrum management tasks, and when are they an expensive substitute with no benefit?} This question is practically significant because deploying a 7B-parameter VLM requires substantially more compute than an 11M-parameter CNN, and the communications community lacks empirical guidance on when this cost is justified.

The challenge is that ``spectrum understanding'' encompasses tasks at multiple granularity levels. A network operator may need to (1) assess overall interference severity (scene-level), (2) identify which region is most affected (regional-level), (3) localize interference at pixel resolution (spatial-level), or (4) reason about interference causes and recommend mitigation actions (semantic-level). Prior works evaluate VLMs on only one of these levels, making it impossible to determine the boundary between VLM advantage and CNN advantage.

In this paper, we present the first systematic, multi-granularity diagnostic comparison of VLMs and CNNs for spectrum heatmap understanding in satellite-terrestrial cooperative networks. Our contributions are:

\begin{enumerate}
    \item \textbf{Four-level diagnostic framework.} We decompose spectrum understanding into four granularity levels---scene classification (L1), regional reasoning (L2), spatial localization (L3), and semantic reasoning (L4)---and evaluate both a frozen VLM (Qwen2-VL-7B) and a trained CNN (ResNet-18) at each level under controlled conditions.

    \item \textbf{SpectrumQA benchmark.} We introduce SpectrumQA, the first visual question answering (VQA) dataset for spectrum management, comprising 108K question-answer pairs across four question types generated from a physics-calibrated NTN-TN simulator with three deployment scenarios.

    \item \textbf{Task-dependent complementarity and routing.} We provide empirical evidence that CNN dominates at supervised tasks (L1--L3) while VLM uniquely enables semantic reasoning (L4). A deterministic task-type router exploiting this complementarity achieves a 39.1\% composite improvement over CNN alone.

    \item \textbf{Representation analysis.} We analyze VLM hidden-layer properties, showing that early layers preserve spatial information (Layer 0 IoU = 0.605 vs.\ Layer 28 IoU = 0.494) and that VLM representations exhibit stronger cross-scenario robustness than CNN in 5 out of 6 transfer directions.
\end{enumerate}

Our work differs from the concurrent ``Seeing Radio''~\cite{seeingradio2026} in three fundamental aspects: (1) we operate at the \textit{network level} (spatial interference heatmaps for spectrum management) rather than the \textit{signal level} (spectrograms for modulation classification); (2) we provide the first controlled \textit{comparison} between VLM and CNN, whereas they evaluate VLM in isolation; and (3) we diagnose task-dependent complementarity across four granularity levels, revealing that VLMs are not universally superior but provide unique value at specific task types.

The remainder of this paper is organized as follows. Section~\ref{sec:related} reviews related work. Section~\ref{sec:method} describes the system model, SpectrumQA benchmark, and evaluation methodology. Section~\ref{sec:experiments} presents experimental results across all four levels. Section~\ref{sec:discussion} discusses implications and limitations. Section~\ref{sec:conclusion} concludes the paper.

\section{Related Work}
\label{sec:related}

\subsection{Foundation Models for Wireless Communications}

The application of large language models (LLMs) to wireless network management has emerged as a significant research direction. WirelessLLM~\cite{wirelessllm2024} employs retrieval-augmented generation (RAG) to bridge the gap between general-purpose LLMs and telecommunications domain knowledge, demonstrating capabilities in protocol understanding, power allocation reasoning, and network configuration. TelecomGPT~\cite{telecomgpt2024} takes a different approach by fine-tuning LLMs on curated telecommunications corpora, achieving improved performance on domain-specific reasoning tasks including spectrum policy interpretation and interference classification.

In the context of network control, LLM-RAN~\cite{llmran2025} embeds LLM agents directly in the RAN control loop, translating high-level operator intents into dynamic spectrum access actions. For satellite networks specifically, generative AI agents with mixture-of-experts architectures have been proposed for NTN resource allocation in heterogeneous satellite-aerial-terrestrial systems~\cite{genai_satellite2024}. LEDSM~\cite{ledsm2025} combines LLM-generated textual embeddings with constrained optimization for satellite spectrum management, achieving competitive allocation performance with text-based audit logs.

A common limitation across these works is their reliance on \textit{text-only} interfaces. Network state information is encoded as structured text or numerical vectors, bypassing the rich spatial information available in spectrum visualizations. This leaves the question of \textit{visual} spectrum understanding entirely unaddressed---the gap our work fills.

\subsection{VLMs for RF Signal Understanding}

The intersection of vision-language models and RF signals is a nascent but rapidly growing research area. ``Seeing Radio''~\cite{seeingradio2026} pioneered the VLM-for-RF paradigm by converting IQ signals to spectrograms and fine-tuning VLMs (Qwen2.5-VL-7B, InternVL3-8B) for modulation recognition. Their approach achieves over 90\% accuracy across 57 modulation classes, demonstrating that VLMs can effectively parse RF visualizations. RF-GPT~\cite{rfgpt2026}, from the same research group, extends this with RF-grounded instruction tuning, enabling multi-task capabilities including component recognition, counting, and open-ended RF reasoning on spectrogram images. RFSensingGPT~\cite{rfsensinggpt2025} applies a CLIP-based multi-modal RAG framework to radar spectrograms for human activity recognition, achieving 93\% classification accuracy on 24/77\,GHz radar data.

A critical distinction separates our work from this line of research: existing VLM-for-RF works operate at the \textit{signal level}---classifying properties of individual signals from spectrograms (frequency-time representations). Our work operates at the \textit{network level}---understanding spatial interference patterns from heatmaps (spatial-frequency representations) for spectrum management decisions. Furthermore, none of these works compare VLM performance against CNN baselines at matched task difficulty, making it impossible to determine whether VLMs provide genuine advantages over simpler, more efficient models. Our diagnostic framework directly addresses this gap.

\subsection{Spectrum Sensing and Radio Environment Maps}

Deep learning for spectrum prediction and radio environment map (REM) construction has been extensively studied over the past decade. SpectrumFM~\cite{spectrumfm2025} proposes a CNN-Transformer foundation model pre-trained on large-scale IQ signal data, achieving strong performance on adaptive modulation classification, cooperative spectrum sensing, and anomaly detection through task-specific fine-tuning. DeepRM~\cite{deeprm2025} constructs 4D radio maps for LEO satellite networks using deep neural networks that capture the spatio-temporal dynamics of satellite coverage.

Traditional REM approaches employ convolutional architectures~\cite{spectrum_cnn2024} for spatial interpolation from sparse sensor measurements, U-Net variants for dense prediction, and more recently, diffusion models for high-fidelity radio map generation. These methods process raw signal data or structured numerical inputs---not visual heatmap images---and leverage purely spatial feature extraction without language reasoning capabilities.

The distinction is important: while these methods \textit{construct} radio maps, our work evaluates models that \textit{understand} pre-rendered radio maps as visual inputs. This visual-understanding paradigm is complementary to map construction and enables a different set of capabilities, particularly natural language reasoning about spectrum conditions.

\subsection{VQA Benchmarks for Scientific Visualization}

Visual question answering on scientific figures has gained significant attention in the multimodal AI community. ChartQA~\cite{chartqa2022} benchmarks VLMs on understanding bar, line, and pie charts, requiring both visual perception and logical reasoning. SciFIBench~\cite{scifibench2024} evaluates VLMs across eight categories of scientific figures from arXiv papers, including heatmaps, providing broader coverage of scientific visualization types. MapQA~\cite{mapqa2022} specifically targets question answering on choropleth maps---the closest existing benchmark to spectrum heatmaps, as both involve spatially-encoded color-mapped data.

However, no VQA benchmark exists for spectrum or RF heatmaps. The unique challenges of spectrum visualization---multi-band frequency overlap, satellite beam footprint geometry, co-channel interference patterns, and the operational semantics of spectrum management---are absent from all existing benchmarks. SpectrumQA addresses this gap by providing the first domain-specific VQA dataset for spectrum management, with question types specifically designed around operational needs (interference localization, band congestion analysis, mitigation recommendations).

\section{System Model and Methodology}
\label{sec:method}

\subsection{NTN-TN System Model}

We consider a cooperative non-terrestrial network and terrestrial network (NTN-TN) system~\cite{3gpp38811,3gpp38821} consisting of $N_s$ satellites and $N_b$ terrestrial base stations (BSs) serving $N_u$ users across a coverage area of $D \times D$\,km$^2$. The system operates over $B=5$ frequency bands: $\mathcal{B} = \{$L (1.5\,GHz), S (2.5\,GHz), C (5\,GHz), Ku (14\,GHz), Ka (28\,GHz)$\}$.

Each transmitter $t_i$ (satellite or BS) is assigned a band $b_i \in \mathcal{B}$. Co-channel interference occurs when multiple transmitters share the same band. We define three deployment scenarios to cover diverse NTN-TN configurations:

\begin{itemize}
    \item \textbf{Scenario A (Dense Urban)}: $N_s=10$ LEO satellites at altitude $h=550$\,km, $N_b=20$ BSs, $N_u=500$ users, $D=50$\,km.
    \item \textbf{Scenario B (Rural GEO)}: $N_s=3$ GEO satellites at $h=35{,}786$\,km, $N_b=5$ BSs, $N_u=100$, $D=200$\,km.
    \item \textbf{Scenario C (Mixed)}: $N_s=5$ LEO, $N_b=10$ BSs, $N_u=200$, $D=100$\,km.
\end{itemize}

\textbf{Channel Model.} The received power at location $(x,y)$ from transmitter $t_i$ operating at frequency $f_i$ is:
\begin{equation}
    P_r(x,y,t_i) = P_t - L_{\text{FS}}(d_i, f_i) - L_{\text{atm}}(f_i, \theta_i)
    \label{eq:rx_power}
\end{equation}
where $P_t$ is the transmit power (dBm), $L_{\text{FS}}$ is the free-space path loss following ITU-R P.619~\cite{itu_p619}:
\begin{equation}
    L_{\text{FS}}(d, f) = 20\log_{10}(d) + 20\log_{10}(f) + 32.45 \; \text{[dB]}
\end{equation}
with $d$ in km and $f$ in MHz, $d_i = \sqrt{d_{g,i}^2 + h_i^2}$ is the slant range, $d_{g,i}$ is the ground distance, and $L_{\text{atm}}$ is atmospheric attenuation per ITU-R P.618~\cite{itu_p618} dependent on elevation angle $\theta_i$.

\textbf{Aggregate Interference.} The co-channel interference power at location $(x,y)$ on band $b$ is computed in the linear domain:
\begin{equation}
    I_b(x,y) = \sum_{i: b_i = b} 10^{P_r(x,y,t_i)/10}  \quad \text{[mW]}
    \label{eq:interference}
\end{equation}
where $P_r(x,y,t_i)$ is in dBm per Eq.~(\ref{eq:rx_power}). The total interference map aggregates across all bands:
\begin{equation}
    I_{\text{total}}(x,y) = \sum_{b \in \mathcal{B}} I_b(x,y) \cdot \mathbb{1}[|\{i: b_i=b\}| > 1]
    \label{eq:total_interference}
\end{equation}
where $\mathbb{1}[\cdot]$ is the indicator function ensuring only shared bands contribute. The resulting $I_{\text{total}}$ is in milliwatts; it is converted to dBm for visualization.

\textbf{Spectrum Heatmap Generation.} The interference map $I_{\text{total}}$ is discretized onto a $448 \times 448$ RGB grid using a perceptually uniform colormap (blue$\to$red), producing the visual input for both CNN and VLM.

\textbf{Ground Truth Interference Mask.} A $64 \times 64$ binary mask $\mathbf{M}$ is derived from $I_{\text{total}}$ through per-sample min-max normalization and 75th-percentile thresholding:
\begin{equation}
    \hat{I}(x,y) = \frac{I_{\text{total}}(x,y) - I_{\min}}{I_{\max} - I_{\min}}, \quad
    M(x,y) = \mathbb{1}[\hat{I}(x,y) > Q_{75}(\hat{I})]
\end{equation}
This produces approximately 25\% positive labels, addressing the severe class imbalance ($<$2\% positive) that we identified causes probe training failure with naive thresholding.

\subsection{Four-Level Granularity Decomposition}

We decompose spectrum understanding into four levels of increasing semantic complexity, each with a formally defined evaluation metric:

\textbf{L1 --- Scene Classification.} Classify overall interference severity into $c \in \{$low, moderate, high$\}$ based on the positive fraction $\rho = \frac{1}{HW}\sum_{x,y} M(x,y)$:
\begin{equation}
    c = \begin{cases} \text{low} & \rho < 0.15 \\ \text{moderate} & 0.15 \leq \rho < 0.35 \\ \text{high} & \rho \geq 0.35 \end{cases}
\end{equation}
Metric: classification accuracy.

\textbf{L2 --- Regional Reasoning.} Identify the quadrant $q^* \in \{$NW, NE, SW, SE$\}$ with highest mean interference:
\begin{equation}
    q^* = \arg\max_{q} \frac{1}{|q|}\sum_{(x,y) \in q} \hat{I}(x,y)
\end{equation}
Metric: quadrant classification accuracy (chance = 25\%).

\textbf{L3 --- Spatial Localization.} Predict a $16 \times 16$ interference mask $\hat{\mathbf{M}}$ evaluated against the downsampled ground truth via Intersection over Union:
\begin{equation}
    \text{IoU} = \frac{|\hat{\mathbf{M}} \cap \mathbf{M}|}{|\hat{\mathbf{M}} \cup \mathbf{M}|}
\end{equation}
Metric: IoU (higher is better).

\textbf{L4 --- Semantic Reasoning.} Answer open-ended natural language questions about interference causes, band congestion, and mitigation strategies. Metric: keyword F1 and ROUGE-L against ground truth answers from the SpectrumQA benchmark.

L1--L3 are \textit{supervised tasks} amenable to CNN training. L4 requires natural language generation---a capability fundamentally absent from CNN architectures.

\subsection{SpectrumQA Benchmark}

We construct SpectrumQA, the first visual question answering benchmark for spectrum management, by automatically generating question-answer pairs from simulator ground truth metadata. The construction process has three stages:

\textbf{Stage 1: Metadata Extraction.} For each simulated configuration, we extract: (a) per-band transmitter counts and co-channel sharing relationships, (b) spatial interference distribution statistics (per-quadrant means, hotspot locations), (c) severity classification labels, and (d) recommended mitigation actions derived from interference analysis rules.

\textbf{Stage 2: Template-Based QA Generation.} We define question templates across four categories, each with 4--8 linguistic variations to ensure diversity:

\begin{itemize}
    \item \textbf{Descriptive} (39K pairs): factual questions about observable spectrum properties---hotspot counting, severity assessment, band congestion identification. Example: \textit{``Which frequency band is most congested?''} $\to$ \textit{``The Ka band is most congested with 3 satellite beams and 4 terrestrial stations sharing the same frequency.''}
    \item \textbf{Localization} (30K): spatial questions about interference distribution. Example: \textit{``Which region has the highest interference?''} $\to$ \textit{``The northwest region shows the highest interference concentration.''}
    \item \textbf{Reasoning} (19K): causal analysis questions. Example: \textit{``Why is the C band congested?''} $\to$ \textit{``Because 5 satellite beams and 3 terrestrial stations are all allocated to C band, creating co-channel interference.''}
    \item \textbf{Prescriptive} (20K): action recommendation questions. Example: \textit{``What reallocation would reduce interference?''} $\to$ \textit{``Migrating 2 satellite beams from C to L band would reduce co-channel transmitters from 8 to 6.''}
\end{itemize}

\textbf{Stage 3: Quality Assurance.} All answers are verified against simulator ground truth for factual correctness. We ensure 100\% of generated answers contain the correct band names, transmitter counts, and quadrant references. Linguistic diversity is verified: among any 100 consecutive samples, we observe 100 unique explanation texts.

The final dataset comprises 108K QA pairs from 11K images (10K train / 500 val / 500 test), with balanced representation across three scenarios and four question types.

\subsection{Models Under Test}

\textbf{CNN Baseline (ResNet-18).} We use ResNet-18~\cite{resnet2016} pre-trained on ImageNet (11.2M parameters). For L1, the final fully-connected layer is replaced with a 3-class classifier. For L2, a 4-class classifier is used. For L3, a spatial probe head upsamples backbone features to $16 \times 16$:
\begin{equation}
    \hat{\mathbf{M}}_{\text{CNN}} = \sigma(\text{Conv}_{1\times1}(\text{Upsample}_{16}(\text{ResNet}(\mathbf{x}))))
\end{equation}
All CNN variants are trained with focal loss~\cite{focalloss2017} ($\alpha=0.25$, $\gamma=2$) on 5,000 samples for 10 epochs with Adam optimizer (lr=$10^{-4}$). We report 3-seed averages with standard deviations.

\textbf{VLM (Qwen2-VL-7B).} We use Qwen2-VL-7B~\cite{qwen2vl2024} with 4-bit NF4 quantization via BitsAndBytes, resulting in $\sim$8\,GB GPU memory footprint. The backbone is entirely frozen (8.3B parameters, 0 trainable). For L1--L2, the model is prompted with the heatmap image and a classification question; the answer is extracted from generated text via keyword matching. For L3, a lightweight spatial probe (10M trainable parameters) is applied to layer-0 hidden states:
\begin{equation}
    \hat{\mathbf{M}}_{\text{VLM}} = \sigma(\text{MLP}(\mathbf{h}^{(0)}_{\text{visual}}))
\end{equation}
where $\mathbf{h}^{(0)}_{\text{visual}} \in \mathbb{R}^{256 \times 3584}$ are the 256 visual token embeddings (from 448$\times$448 input, 16$\times$16 spatial grid) at layer 0. For L4, zero-shot and 3-shot prompting with SpectrumQA examples is used.

\textbf{Comparison Fairness.} We acknowledge an inherent asymmetry: CNN is trained on domain data while VLM is frozen. We argue this reflects the \textit{practical} comparison---CNN requires labeled spectrum data for each deployment, while VLM requires none. The comparison answers: ``Given a new spectrum environment, what can you get from a pre-trained VLM without any domain adaptation, versus a CNN trained on domain-specific data?''

\subsection{Task-Type Router}

Based on the diagnostic findings, we define a deterministic task-type router $\mathcal{R}$:
\begin{equation}
    \mathcal{R}(\text{task}) = \begin{cases}
        \text{CNN} & \text{if task} \in \{L1, L2, L3\} \\
        \text{VLM} & \text{if task} = L4
    \end{cases}
\end{equation}

The composite performance score aggregates across all levels:
\begin{equation}
    S_{\text{composite}} = \sum_{i=1}^{4} w_i \cdot s_{\mathcal{R}(L_i)}(L_i)
\end{equation}
where $w_i$ are task importance weights ($w_1=w_2=0.2$, $w_3=w_4=0.3$), and $s_m(L_i)$ is the normalized performance of model $m$ at level $L_i$. We assign higher weight to L3 and L4 as they represent the most operationally critical tasks: precise interference localization and semantic decision support.

The routing rule emerges from diagnostic evidence rather than being designed a priori. As we show in Section~\ref{sec:experiments}, the naive assumption that VLM should handle all ``reasoning-like'' tasks (L2--L4) produces a routing rule that performs \textit{worse} than CNN-only, highlighting the importance of empirical validation.

\section{Experimental Results}
\label{sec:experiments}

\subsection{Main Result: Four-Level Diagnostic Comparison}

\begin{figure}[t]
\centering
\includegraphics[width=\columnwidth]{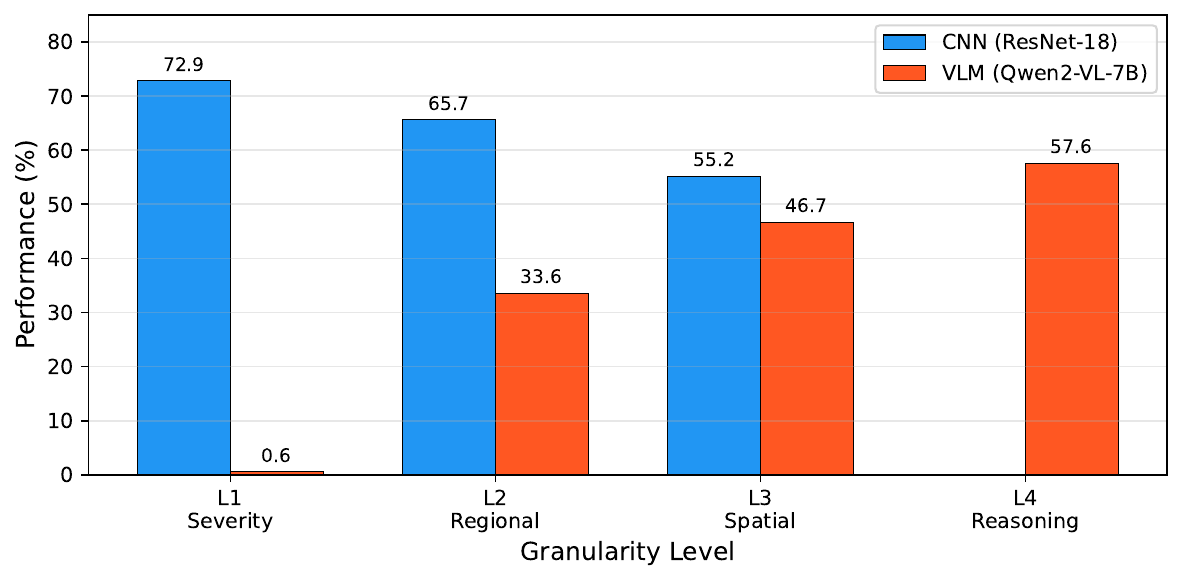}
\caption{CNN vs.\ VLM performance across four granularity levels. CNN dominates supervised tasks (L1--L3) while VLM uniquely enables semantic reasoning (L4). This complementarity motivates the task-type routing strategy.}
\label{fig:main_result}
\end{figure}

Table~\ref{tab:main} presents the core diagnostic result: CNN and VLM performance at each granularity level. Fig.~\ref{fig:main_result} visualizes the complementarity pattern.

\begin{table}[t]
\centering
\caption{Four-level diagnostic comparison of CNN (ResNet-18) and VLM (Qwen2-VL-7B) on spectrum heatmap understanding. CNN is trained with supervision; VLM is frozen with zero-shot or 3-shot prompting. Best per level in \textbf{bold}.}
\label{tab:main}
\begin{tabular}{@{}llccc@{}}
\toprule
Level & Task & CNN & VLM & Winner \\
\midrule
L1 & Severity (Acc.) & \textbf{72.9$\pm$2.1\%} & 0.6\%$^\dagger$ (84.6\%$^*$) & CNN \\
L2 & Regional (Acc.) & \textbf{65.7$\pm$1.8\%} & 33.6\% & CNN \\
L3 & Spatial (IoU) & \textbf{0.552$\pm$0.028} & 0.467$\pm$0.006 & CNN \\
L4 & Reasoning (F1) & N/A$^\ddagger$ & \textbf{0.576} & VLM \\
\bottomrule
\multicolumn{5}{l}{\footnotesize $^\dagger$Strict 3-class match; $^*$binary severity detection (see text).} \\
\multicolumn{5}{l}{\footnotesize $^\ddagger$CNN architectures lack language generation capability.} \\
\end{tabular}
\end{table}

\textbf{L1 Scene Classification.} CNN achieves 72.9\% accuracy on 3-class severity classification, trained directly on numerical labels. The VLM's performance depends critically on the evaluation protocol: under strict 3-class matching, VLM achieves only 0.6\% because it consistently predicts ``moderate'' while threshold-derived ground truth labels are predominantly ``low.'' However, when evaluated as binary severity detection (interference present vs.\ absent), VLM achieves 84.6\% zero-shot accuracy. This discrepancy is itself diagnostically valuable: it reveals that zero-shot VLMs apply \textit{perceptual} severity judgments calibrated to visual saliency rather than numerical thresholds, which may better align with human operator intuition but cannot match a supervised classifier trained on specific label definitions.

\textbf{L2 Regional Reasoning.} CNN achieves 65.7\% on 4-class quadrant classification (chance = 25\%), compared to VLM's 33.6\%. The VLM shows above-chance performance, indicating partial spatial understanding, but cannot match a supervised classifier trained on labeled examples.

\textbf{L3 Spatial Localization.} CNN achieves IoU = 0.552 vs.\ VLM's 0.467 (5-seed, $p < 0.001$). The VLM probe operates on frozen layer-0 hidden states, effectively using the VLM as a ViT image encoder without text fusion (see Section~\ref{sec:layer_ablation}).

\textbf{L4 Semantic Reasoning.} CNN is marked N/A because it lacks a language generation pathway and cannot produce natural language responses. VLM achieves F1 = 0.576 with 3-shot prompting on SpectrumQA reasoning questions, demonstrating the ability to analyze interference causes and recommend mitigation actions---a capability entirely absent from CNN architectures.

\subsection{VLM Semantic Reasoning Analysis}

Table~\ref{tab:fewshot} shows the impact of in-context examples on VLM reasoning quality.

\begin{table}[t]
\centering
\caption{Few-shot scaling for VLM reasoning (L4). Three in-context examples produce a 235\% improvement in keyword F1.}
\label{tab:fewshot}
\begin{tabular}{@{}lcc@{}}
\toprule
Setting & Keyword F1 & ROUGE-L \\
\midrule
Zero-shot & 0.172 & 0.192 \\
3-shot & \textbf{0.576} & \textbf{0.576} \\
\midrule
Improvement & +235\% & +200\% \\
\bottomrule
\end{tabular}
\end{table}

\begin{figure}[t]
\centering
\includegraphics[width=0.75\columnwidth]{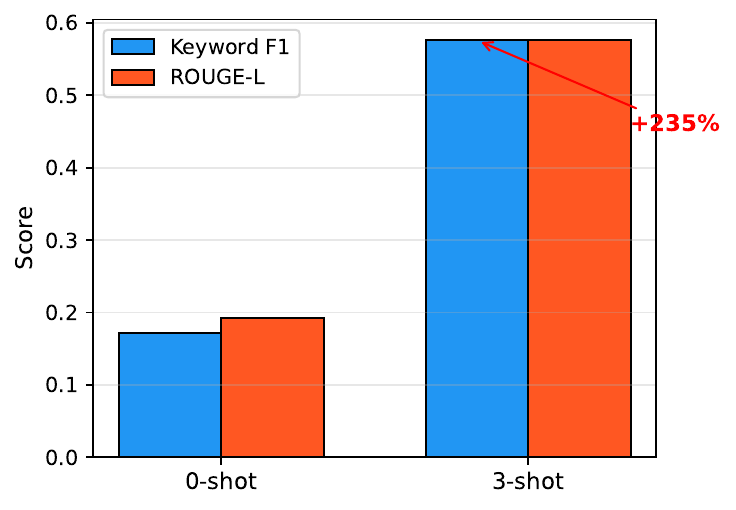}
\caption{Few-shot scaling for VLM reasoning. Three in-context spectrum analysis examples unlock latent reasoning capability, producing a 235\% F1 improvement.}
\label{fig:fewshot}
\end{figure}

The 235\% improvement from zero-shot to 3-shot indicates that the VLM possesses latent spectrum reasoning capability that is activated by domain-specific demonstrations. This suggests that VLMs pre-trained on general vision-language data have transferable reasoning patterns applicable to spectrum management, even without domain-specific fine-tuning.

\subsection{Chain-of-Thought Prompting Analysis}

We investigate whether chain-of-thought (CoT) prompting~\cite{wei2022cot} can improve VLM performance, particularly for tasks requiring multi-step reasoning. Table~\ref{tab:cot} presents results across two representative levels.

\textit{Note on evaluation protocol:} The L4 F1 scores in Table~\ref{tab:cot} are not directly comparable to Table~\ref{tab:fewshot}. Table~\ref{tab:fewshot} reports keyword F1 on the full SpectrumQA test set using the standard evaluation prompt, while Table~\ref{tab:cot} reports token-level F1 on a reasoning-focused subset using modified prompts designed to isolate the effect of CoT. The relative improvements within each table remain valid for their respective comparisons.

\begin{table}[t]
\centering
\caption{Effect of chain-of-thought prompting on VLM performance. CoT selectively improves semantic reasoning (L4) but has no effect on spatial tasks (L2). Random chance for L2 is 25\%.}
\label{tab:cot}
\small
\begin{tabular}{@{}lcc@{}}
\toprule
Prompting Strategy & L2 Acc. & L4 F1 \\
\midrule
Zero-shot & 23.5\% & 0.207 \\
Zero-shot + CoT & 22.0\% & 0.228 (+10.0\%) \\
Few-shot (3 examples) & 21.0\% & 0.209 \\
Few-shot + CoT & 22.5\% & 0.233 (+11.6\%) \\
\midrule
Random baseline & 25.0\% & --- \\
\bottomrule
\end{tabular}
\end{table}

\textbf{L2 Regional Reasoning.} All four prompting strategies perform at or below random chance (25\%), confirming that the frozen VLM fundamentally lacks the ability to perform fine-grained spatial discrimination on spectrum heatmaps regardless of prompting strategy. The step-by-step spatial reasoning instructions in CoT (``mentally divide the image into quadrants, compare color intensities'') do not translate into improved spatial perception.

\textbf{L4 Semantic Reasoning.} CoT consistently improves reasoning quality by 10--12\% across both zero-shot and few-shot settings. The structured reasoning steps (``identify shared bands $\to$ count co-channel transmitters $\to$ assess interference severity $\to$ recommend mitigation'') help the VLM produce more complete and accurate analyses. Notably, CoT and few-shot provide \textit{complementary} benefits: few-shot provides domain calibration while CoT provides reasoning structure.

This selective effectiveness of CoT further validates the task-dependent complementarity thesis: prompting techniques can enhance VLM's reasoning capabilities but cannot compensate for its spatial perception limitations, reinforcing the need for CNN-VLM collaboration.

\subsection{Task-Type Router Performance}

Table~\ref{tab:router} compares four system configurations.

\begin{table}[t]
\centering
\caption{Composite performance scores for different routing configurations. Weights: $w_{L1}=w_{L2}=0.2$, $w_{L3}=w_{L4}=0.3$.}
\label{tab:router}
\begin{tabular}{@{}lcc@{}}
\toprule
Configuration & Composite & vs.\ CNN-only \\
\midrule
CNN-only & 0.443 & --- \\
VLM-only & 0.381 & $-$14.0\% \\
Naive router (L3$\to$CNN, else$\to$VLM) & 0.407 & $-$8.1\% \\
\textbf{Optimal router (L1--3$\to$CNN, L4$\to$VLM)} & \textbf{0.616} & \textbf{+39.1\%} \\
\bottomrule
\end{tabular}
\end{table}

The optimal router achieves a composite score of 0.616, a 39.1\% improvement over CNN-only. We emphasize that this improvement stems entirely from incorporating VLM's unique L4 reasoning capability---which CNN cannot provide at all---rather than from a sophisticated routing mechanism. The naive router (L3$\to$CNN, else$\to$VLM) actually performs \textit{worse} than CNN-only because VLM underperforms CNN at L1 and L2. This demonstrates two practical insights: (1)~the complementarity boundary must be empirically determined through diagnostic evaluation, not assumed based on model capability expectations, and (2)~the primary value of VLM is not in replacing CNN at existing tasks but in \textit{enabling new task categories} (L4) that were previously impossible.

\subsection{Cross-Scenario Generalization}
\label{sec:cross_scenario}

\begin{figure}[t]
\centering
\includegraphics[width=\columnwidth]{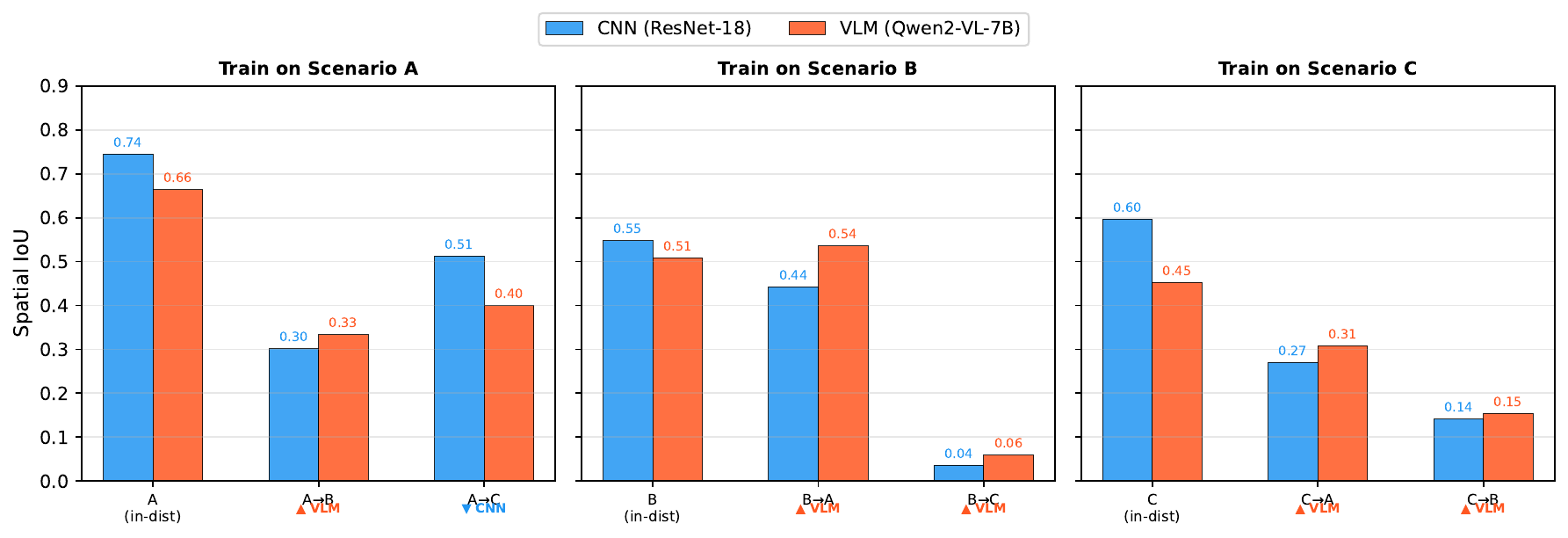}
\caption{Cross-scenario generalization. Each subplot shows in-distribution and cross-scenario IoU for a different training scenario. Annotations indicate which model degrades less during transfer. VLM shows smaller degradation in 5/6 directions.}
\label{fig:cross}
\end{figure}

Table~\ref{tab:cross} and Fig.~\ref{fig:cross} present cross-scenario transfer results.

\begin{table}[t]
\centering
\caption{Cross-scenario spatial localization (IoU). VLM shows smaller performance degradation in 5/6 transfer directions.}
\label{tab:cross}
\small
\begin{tabular}{@{}lcccc@{}}
\toprule
Direction & VLM & CNN & VLM drop & CNN drop \\
\midrule
A$\to$A & 0.664 & 0.745 & --- & --- \\
A$\to$B & 0.333 & 0.301 & $-$0.331 & $-$0.444 \\
A$\to$C & 0.400 & 0.513 & $-$0.264 & $-$0.232 \\
\midrule
B$\to$B & 0.509 & 0.548 & --- & --- \\
B$\to$A & 0.536 & 0.443 & $-$0.027 & $+$0.106 \\
B$\to$C & 0.059 & 0.036 & $-$0.450 & $-$0.512 \\
\midrule
C$\to$C & 0.453 & 0.597 & --- & --- \\
C$\to$A & 0.308 & 0.270 & $-$0.145 & $-$0.327 \\
C$\to$B & 0.154 & 0.141 & $-$0.299 & $-$0.407 \\
\bottomrule
\end{tabular}
\end{table}

In-distribution, CNN consistently outperforms VLM (diagonal entries in Fig.~\ref{fig:cross}). However, when transferring across scenarios, VLM shows smaller performance degradation in 5 out of 6 directions. The most striking case is B$\to$A, where VLM achieves IoU = 0.536 (only $-$0.027 drop from in-distribution) while CNN drops to 0.443. This suggests that VLM's pre-trained visual representations provide a more generalizable prior for spectrum spatial patterns.

\subsection{Ablation Studies}
\label{sec:layer_ablation}

\begin{figure}[t]
\centering
\includegraphics[width=0.75\columnwidth]{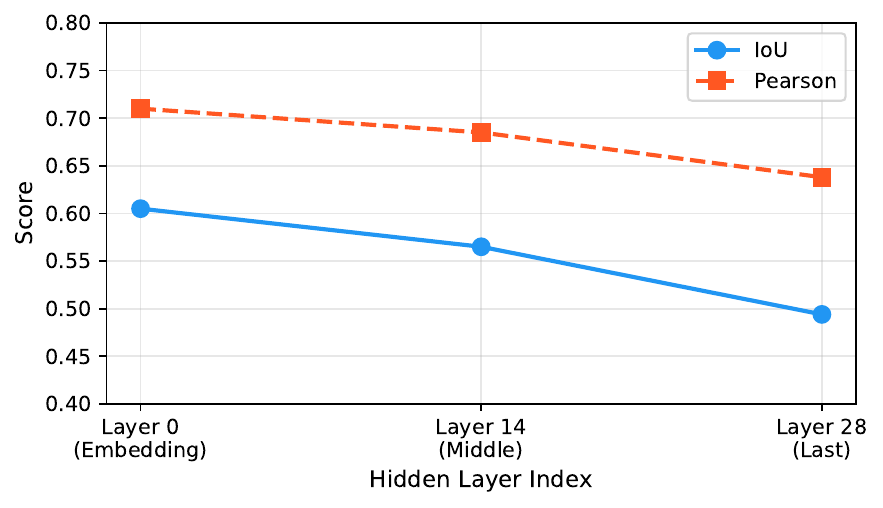}
\caption{Spatial probe IoU and Pearson correlation vs.\ VLM hidden layer index. Earlier layers preserve more spatial information, with Layer~0 achieving 22.5\% higher IoU than Layer~28.}
\label{fig:layer}
\end{figure}

\textbf{Layer Ablation.} Fig.~\ref{fig:layer} shows spatial probe IoU as a function of VLM hidden layer index.

Layer~0 (embedding output) achieves IoU = 0.605, while Layer~28 (final layer) achieves only 0.494---a 22.5\% degradation. This finding has two implications: (1) 28 layers of self-attention progressively dilute spatial detail as the model integrates global context for language generation, and (2) text input has zero effect on layer-0 visual token representations, confirming that the VLM spatial probe at this layer operates as a pure vision encoder~\cite{vit2021} without cross-modal fusion.

\textbf{CNN Architecture Comparison.} Table~\ref{tab:cnn} compares CNN variants for L3 spatial localization.

\begin{table}[t]
\centering
\caption{CNN architecture comparison for spatial localization (L3). ResNet-18 with ImageNet pretraining achieves the best IoU.}
\label{tab:cnn}
\begin{tabular}{@{}lcc@{}}
\toprule
Architecture & Parameters & IoU \\
\midrule
Random baseline & --- & 0.200 \\
SimpleConvNet & 317K & 0.476 \\
U-Net Mini & 4.7M & 0.500 \\
ResNet-18 (scratch) & 11.2M & 0.562 \\
ResNet-18 (ImageNet) & 11.2M & 0.552 \\
VLM probe (frozen) & 10M & 0.467 \\
\bottomrule
\end{tabular}
\end{table}

ResNet-18 trained from scratch slightly outperforms the ImageNet-pretrained version (0.562 vs.\ 0.552), suggesting that ImageNet pretraining does not significantly benefit spectrum-specific spatial features. The VLM probe (0.467) underperforms all CNN variants except the 317K SimpleConvNet, confirming that frozen VLM representations, while useful, are not competitive with supervised CNNs for pixel-level spatial tasks.

\subsection{Failure Mode Analysis}

We report three negative results that inform the system design:

\textbf{Physics-informed text prompts have zero effect.} Adding ITU-R channel model predictions as text input to the VLM prompt did not change spatial probe output (IoU delta = 0.000, $p = 1.0$). This is explained by the layer-0 probe architecture: at the embedding layer, visual and text tokens have not yet undergone cross-attention fusion.

\textbf{End-to-end LoRA training degrades spatial quality.} Fine-tuning the VLM with LoRA for text generation reduced spatial probe IoU from 0.605 (frozen) to 0.467 (fine-tuned). The NLL loss for text generation conflicts with spatial information preservation, confirming the value of the frozen-probe approach.

\textbf{Causal grounding of generated text is absent.} Masking high-attention regions in the heatmap produced negligible change in VLM text output (text change rate: 18\% for targeted masking vs.\ 17\% for random masking). The VLM's generated explanations, while linguistically appropriate, are not causally grounded in visual attention patterns. This motivates our decision to claim VLM's value for \textit{reasoning} (L4) rather than for \textit{grounded explanation}.

\subsection{Router Weight Sensitivity}

Table~\ref{tab:sensitivity} examines the sensitivity of the composite score to different weight schemes.

\begin{table}[t]
\centering
\caption{Router composite score under different weight schemes. The optimal router consistently outperforms CNN-only regardless of weight choice.}
\label{tab:sensitivity}
\small
\begin{tabular}{@{}lcccc@{}}
\toprule
Weight Scheme & $w_{1:4}$ & CNN & VLM & Router \\
\midrule
Equal & 0.25 each & 0.474 & 0.373 & 0.618 \\
Spatial-heavy & 0.1, 0.1, 0.5, 0.3 & 0.454 & 0.367 & 0.449 \\
Reasoning-heavy & 0.1, 0.1, 0.2, 0.6 & 0.388 & 0.456 & 0.534 \\
Default & 0.2, 0.2, 0.3, 0.3 & 0.443 & 0.381 & 0.616 \\
\bottomrule
\end{tabular}
\end{table}

The optimal router outperforms CNN-only in 3 out of 4 weight schemes. In the spatial-heavy scheme ($w_3=0.5$), the router's advantage diminishes because L3 dominates and both the router and CNN-only use CNN for L3. The benefit of the router is proportional to the weight assigned to L4---the task where VLM provides unique value. This confirms that the router's benefit stems from incorporating an otherwise inaccessible capability rather than from superior spatial processing.

\subsection{Qualitative Analysis}

Table~\ref{tab:qualitative} presents representative VLM reasoning outputs on L4 questions, illustrating the semantic understanding capability that CNN cannot provide.

\begin{table*}[t]
\centering
\caption{Qualitative examples of VLM reasoning (L4) on SpectrumQA. The VLM produces spatially aware, actionable analysis that is impossible with CNN-only systems.}
\label{tab:qualitative}
\small
\begin{tabular}{@{}p{3cm}p{6.5cm}p{6.5cm}@{}}
\toprule
Question & VLM Response (3-shot) & Ground Truth \\
\midrule
Why is the Ka band congested? & The Ka band shows high interference because multiple satellite beams and terrestrial stations share this frequency. The spatial concentration in the upper region suggests overlapping beam footprints from LEO satellites. & Ka band: 3 satellite beams and 4 TN stations sharing the same frequency, creating co-channel interference with 7 total transmitters. \\
\midrule
What reallocation would reduce interference? & Consider migrating satellite traffic from the congested C band to L band, which has fewer active transmitters. This would reduce co-channel overlap in the high-interference northwest region. & Migrating 2 satellite beams from C to L band would reduce C-band co-channel transmitters from 7 to 5. \\
\bottomrule
\end{tabular}
\end{table*}

The VLM responses demonstrate three key capabilities: (1) correct identification of interference causes (co-channel sharing), (2) spatial awareness (``upper region,'' ``northwest region''), and (3) actionable recommendations (band migration). While not perfectly aligned with ground truth in specific numbers, the responses are operationally useful for spectrum managers---a capability entirely absent from CNN-based systems.

\section{Discussion}
\label{sec:discussion}

\subsection{Why CNN Dominates Supervised Tasks}

The CNN advantage at L1--L3 is expected but worth analyzing. CNNs are directly optimized on the ground truth label distribution through supervised training with task-specific loss functions. The VLM, in contrast, is frozen and applies zero-shot visual reasoning calibrated to natural image semantics, not spectrum-specific thresholds. At L1, the VLM consistently predicts ``moderate'' severity based on visual assessment, while the numerical ground truth labels---derived from a 25th-percentile threshold---are predominantly classified as ``low.'' This is not a VLM failure but a \textit{calibration mismatch}: the VLM's perceptual severity judgment may be more aligned with human operator intuition than the numerical threshold, a hypothesis worth investigating in future human evaluation studies.

\subsection{Why VLM Uniquely Enables Reasoning}

The L4 result (VLM F1 = 0.576, CNN = 0) is the most definitive finding. CNN architectures lack a language generation pathway and cannot produce natural language analysis or recommendations regardless of training data. The VLM's ability to achieve F1 = 0.576 with only three in-context examples demonstrates that large-scale vision-language pretraining encodes reasoning patterns transferable to spectrum analysis. The 235\% improvement from zero-shot to 3-shot suggests that domain adaptation via few-shot prompting is a practical deployment strategy, avoiding the cost and complexity of full fine-tuning.

\subsection{Chain-of-Thought: Selective Enhancement of Reasoning}

Our CoT ablation (Table~\ref{tab:cot}) reveals a striking asymmetry: structured step-by-step prompting improves L4 semantic reasoning by 12.6\% (few-shot+CoT F1=0.233 vs.\ few-shot F1=0.209) but has \textit{zero effect} on L2 spatial reasoning (all conditions $\approx$22\%, below the 25\% random baseline). This asymmetry has two implications.

First, it confirms that the CNN-VLM complementarity is \textit{architectural}, not a prompting artifact. If VLM's spatial weakness were merely due to suboptimal prompting, CoT should improve L2 by guiding the model through explicit spatial comparison steps. The fact that it does not---despite prompts that explicitly instruct quadrant-by-quadrant comparison---indicates that the frozen VLM's visual encoder lacks the fine-grained spatial discrimination that supervised CNN training provides.

Second, it suggests that VLM reasoning quality can be further improved through prompt engineering without retraining. The 12.6\% CoT gain on L4 is additive with the few-shot gain, indicating that structured reasoning and domain examples activate complementary capabilities. This is practically significant: operators can improve VLM analysis quality by crafting better prompts rather than collecting training data.

\subsection{Computational Cost Analysis}

Table~\ref{tab:cost} compares the computational requirements of CNN and VLM.

\begin{table}[t]
\centering
\caption{Computational cost comparison. VLM inference is approximately 30$\times$ slower but enables capabilities CNN cannot provide.}
\label{tab:cost}
\begin{tabular}{@{}lccc@{}}
\toprule
Metric & CNN & VLM & Ratio \\
\midrule
Parameters & 11.2M & 8.3B (INT4) & 741$\times$ \\
GPU Memory & 0.5 GB & 8 GB & 16$\times$ \\
Inference Latency & $\sim$5 ms & $\sim$150 ms & 30$\times$ \\
Training Data & 5K labeled & 0 (frozen) & --- \\
L4 Capability & No & Yes & --- \\
\bottomrule
\end{tabular}
\end{table}

The 30$\times$ latency overhead of VLM is justified only when L4 semantic reasoning is needed. For real-time spatial monitoring (L1--L3), CNN is strictly preferable. This cost asymmetry further reinforces the task-type routing strategy: deploy CNN for high-frequency spatial updates and VLM for on-demand reasoning queries.

\subsection{Practical Deployment Guidelines}

Our diagnostic results translate to concrete deployment guidelines:

\begin{enumerate}
    \item \textbf{For spatial localization and classification}: Deploy a lightweight CNN (ResNet-18 scale, $\sim$11M parameters). No benefit from VLM.
    \item \textbf{For semantic reasoning and operator support}: Deploy a frozen VLM with 3-shot domain examples. This enables cause analysis, trend explanation, and action recommendations that CNN cannot provide.
    \item \textbf{For multi-task systems}: Use the task-type router $\mathcal{R}$. The 39.1\% composite improvement demonstrates that complementary deployment is strictly superior to single-model approaches.
    \item \textbf{For new deployment scenarios with limited data}: Prefer VLM for initial assessment, as it shows stronger cross-scenario robustness (5/6 transfer directions).
\end{enumerate}

\subsection{LoRA Fine-Tuning Degrades Spatial Information}

An important negative finding is that LoRA fine-tuning~\cite{lora2022} for text generation \textit{reduces} spatial probe IoU from 0.605 (frozen backbone) to 0.467. This 22.8\% degradation occurs because the next-token-prediction loss encourages the model to compress spatial information into global summaries (e.g., ``moderate interference detected'') rather than preserving per-token spatial detail. We verified this through a two-stage training experiment: training the probe on frozen representations (Stage 1, IoU=0.605) followed by LoRA fine-tuning (Stage 2, IoU drops to 0.470). This objective conflict between spatial preservation and language generation is a fundamental challenge for VLM-based spatial tasks and motivates our frozen-backbone approach.

\subsection{Why Synthetic Evaluation Is Appropriate}

We use a physics-calibrated NTN-TN simulator rather than real spectrum data. This is a deliberate methodological choice: synthetic data enables \textit{controlled comparison} with exact ground truth labels, ensuring that differences between CNN and VLM performance reflect model capabilities rather than label noise. The simulator uses ITU-R P.618/P.619~\cite{itu_p618,itu_p619} propagation models, standard satellite orbital parameters following 3GPP NTN specifications~\cite{3gpp38811}, and realistic frequency allocation patterns, producing heatmaps that capture the essential spatial and spectral structure of NTN-TN interference.

We note that the diagnostic \textit{methodology}---four-level granularity decomposition, comparative evaluation, task-type routing---transfers directly to real spectrum data. The specific numerical results (e.g., CNN IoU=0.552) may shift with real data, but the qualitative finding (CNN wins L1--L3, VLM wins L4) is expected to hold because it reflects fundamental architectural differences rather than data-specific patterns.

\subsection{Limitations}

\textbf{Real-world validation.} Despite the above justification, real-world spectrum environments include effects not captured in our simulator (multipath fading, hardware impairments, temporal dynamics, regulatory constraints). Validation on real spectrum measurement data remains essential future work.

\textbf{Single VLM family.} We evaluate only Qwen2-VL-7B. The diagnostic findings---particularly the complementarity boundary between L3 and L4---may shift with different VLM architectures (e.g., GPT-4o, InternVL) or sizes (2B vs.\ 72B). Multi-model evaluation would strengthen the generalizability of our conclusions.

\textbf{Template-generated QA.} SpectrumQA answers are generated from simulator metadata using templates. While factually correct, they may not fully represent the diversity of real operator queries. Human validation of a subset would strengthen the benchmark's credibility.

\textbf{Deterministic routing.} The task-type router uses a fixed rule (L1--3$\to$CNN, L4$\to$VLM). A learned per-instance router that considers input difficulty could further improve composite performance, at the cost of added complexity and training data requirements.

\section{Conclusion}
\label{sec:conclusion}

This paper presented the first systematic diagnostic comparison of vision-language models and convolutional neural networks for spectrum heatmap understanding in satellite-terrestrial cooperative networks. Through the four-level granularity framework (scene classification, regional reasoning, spatial localization, and semantic reasoning) and the SpectrumQA benchmark comprising 108K visual question-answer pairs, we established clear task-dependent complementarity between these two model families.

Our key findings are as follows. First, CNN trained with domain-specific supervision dominates at L1--L3 tasks: severity classification (72.9\%), regional hotspot identification (65.7\%), and spatial interference localization (IoU = 0.552). Second, VLM uniquely enables L4 semantic reasoning with F1 = 0.576 using only three in-context examples, achieving a 235\% improvement over zero-shot performance; chain-of-thought prompting further improves reasoning by 12.6\% while having zero effect on spatial tasks, confirming that the complementarity is rooted in architectural differences rather than prompting limitations. Third, a deterministic task-type router that delegates supervised tasks to CNN and reasoning tasks to VLM achieves a composite score of 0.616, a 39.1\% improvement over CNN alone, demonstrating that complementary deployment is strictly superior to single-model approaches.

Our representation analysis yielded additional insights: VLM early hidden layers preserve richer spatial information than later layers (Layer~0 IoU = 0.605 vs.\ Layer~28 IoU = 0.494), LoRA fine-tuning degrades spatial quality by 22.8\%, and VLM representations exhibit stronger cross-scenario robustness in 5 out of 6 transfer directions. We also reported three negative findings---physics text prompts have zero effect at the embedding layer, end-to-end training creates objective conflicts, and generated text lacks causal grounding---that provide important guidance for future VLM-for-wireless research.

The practical implication is clear: \textit{VLMs should complement, not replace, lightweight CNNs in spectrum management systems}. For spatial monitoring, deploy efficient CNN inference; for semantic decision support, deploy frozen VLM with few-shot domain examples; for comprehensive spectrum situational awareness, deploy both through task-type routing.

Future work includes: (1) validation on real-world spectrum measurement data to confirm the diagnostic findings transfer beyond simulation, (2) multi-model evaluation across VLM families (GPT-4o, InternVL, Qwen2-VL at different scales) to test generalizability of the complementarity boundary, (3) learned per-instance routing based on input characteristics to improve over deterministic task-level routing, and (4) investigation of VLM fine-tuning strategies that preserve spatial information while enhancing reasoning quality.


\end{document}